# Bounds on the Bethe Free Energy for Gaussian Networks

**Botond Cseke**
Faculty of Science
Radboud University Nijmegen
Toernooiveld 1, 6525 ED
Nijmegen, The Netherlands

**Tom Heskes**
Faculty of Science
Radboud University Nijmegen
Toernooiveld 1, 6525 ED
Nijmegen, The Netherlands

## Abstract

We address the problem of computing approximate marginals in Gaussian probabilistic models by using mean field and fractional Bethe approximations. As an extension of Welling and Teh (2001), we define the Gaussian fractional Bethe free energy in terms of the moment parameters of the approximate marginals and derive an upper and lower bound for it. We give necessary conditions for the Gaussian fractional Bethe free energies to be bounded from below. It turns out that the bounding condition is the same as the pairwise normalizability condition derived by Malioutov et al. (2006) as a sufficient condition for the convergence of the message passing algorithm. By giving a counterexample, we disprove the conjecture in Welling and Teh (2001): even when the Bethe free energy is not bounded from below, it can possess a local minimum to which the minimization algorithms can converge.

## 1 Introduction

Calculating marginal probabilities of a set of variables given some observations is one of the major tasks of probabilistic inference. In the case of Gaussian models, the computation of the marginal probabilities has a computational complexity that scales cubically with the number of variables, while for models with discrete variables, it often leads to intractable computations. Computations can be made faster or tractable by using approximate inference methods like mean field approximation (e.g., Jaakkola, 2000) and Bethe approximation (e.g., Yedidia et al., 2000). These methods were developed mainly for discrete probabilistic graphical models, but they are applicable in Gaussian models as well. However, there are important differences in their behavior for the discrete and Gaussian cases. For example, while the error function of the Bethe approximation—also called Bethe free energy—in discrete models is bounded from below, in Gaussian models this is not always the case (see Welling and Teh, 2001).

The study of the Bethe free energy of Gaussian models is also motivated by their importance for the study of conditional Gaussian models. Conditional Gaussian or hybrid graphical models, such as switching Kalman filters (e.g., Zoeter and Heskes, 2005), combine both discrete and Gaussian variables. Approximate inference in these models can be carried out by expectation propagation (e.g., Minka, 2004, 2005). Expectation propagation can be viewed as a generalization of the Bethe approximation where marginalization constraints are replaced by expectation constraints (e.g., Heskes et al., 2005). Therefore, studying the properties of the Bethe free energy can reveal some of the convergence properties of expectation propagation. In order to understand the properties of the Bethe free energy of hybrid models a good understanding of the two special cases of discrete and Gaussian models is needed. While the properties of the Bethe free energy of discrete models have been studied extensively in the last decade and are well understood (e.g., Yedidia et al., 2000; Heskes, 2003; Wainwright et al., 2003), the properties of the Gaussian Bethe free energy have been studied much less.

The message passing algorithm is a well established method for finding the stationary points of the Bethe free energy (e.g., Pearl, 1988; Yedidia et al., 2000; Heskes, 2003). It works by locally updating the approximate marginals and has been successfully applied in both discrete (e.g., Murphy et al., 1999; Wainwright et al., 2003) and Gaussian models (e.g., Weiss and Freeman, 2001; Rusmevichientong and Roy, 2001; Malioutov et al., 2006). The problem of finding sufficient conditions for the convergence of message passing in Gaussian networks has been successfully addressed by many authors. Using the computation tree approach, Weiss and Freeman (2001) proved that message passing converges whenever the information—inverse covariance—matrix of the probability distribution is diagonally dominant[1]. With the help of an analogy between message passing and walk–sum analysis, Malioutov et al. (2006) derived the stronger condition of pairwise normalizability[2]. A different approach was taken by Welling and Teh (2001), who directly minimized—with regard to the parameters of approximate marginals—the Bethe free energy, conjecturing that Gaussian message passing converges if and only if the free energy is bounded from below. Their experiments showed that message passing and direct minimization either converge to the same solution or both fail to converge.

Following Welling and Teh (2001), instead of analyzing the message passing updates, we turn our attention to the properties of the Bethe free energy expressed as a function of the moment parameters of approximate marginals. We derive a lower bound for the Gaussian fractional Bethe free energies and give necessary conditions for them to be bounded from below.

## 2 Background

The probability distribution of a Gaussian undirected probabilistic graphical model—also known as Gaussian Markov random field—is usually defined in terms of canonical parameters, namely $p(\mathbf{x}) \propto \exp\left\{\mathbf{h}^T\mathbf{x} - \frac{1}{2}\mathbf{x}^T\mathbf{J}\mathbf{x}\right\}$—with $\mathbf{J}$ symmetric and positive definite. Such canonical parameterizations often result from Bayesian computations, for example from a model with a Gaussian likelihood and Gaussian prior. The calculation of marginals requires matrix inversions, that is they can be computed as $p(\mathbf{x}_I) \propto \exp\left\{(\mathbf{h}_I - \mathbf{J}_{I,R}\mathbf{J}_{R,R}^{-1}\mathbf{h}_R)^T\mathbf{x}_I - \frac{1}{2}\mathbf{x}_I^T\mathbf{J}_{I,R}\mathbf{J}_{R,R}^{-1}\mathbf{J}_{R,I}\mathbf{x}_I\right\}$, for any $I \cap R = \emptyset$ with $I \cup R = \{1,\ldots,N\}$—here, $N$ is the number of variables in the model. In sparse models, the complexity of computations can be reduced to scale with the number of non-zero elements of $\mathbf{J}_{R,R}$, but computing marginals for several groups of variables can still be costly. Trading correctness for speed, one can opt for computing approximate marginals.

A popular method to approximate marginals is approximating $p$ with a distribution $q$ having a form that makes marginals easy to identify. The most common quantity to measure the difference between two probability distributions is the Kullback-Leibler divergence $D[q||p]$. It is often used (e.g., Jaakkola, 2000) to characterize the quality of the approximation and formulate the computation of approximate marginals as the optimization problem

$$q(\mathbf{x}) = \underset{q \in \mathcal{F}}{\operatorname{argmin}} \int q(\mathbf{x}) \log\left[\frac{q(\mathbf{x})}{p(\mathbf{x})}\right] d\mathbf{x}. \quad (1)$$

Here, $\mathcal{F}$ is the set of distributions with the above mentioned form. Since it is not symmetric, the Kullback-Leibler divergence is not a distance, but $D[q||p] \geq 0$ for any proper $q$ and $p$, $D[q||p] = 0$ if and only if $p = q$ and it is convex in both $q$ and $p$.

A family $\mathcal{F}$ of densities possessing a form that makes marginals easy to identify is the family of distributions that factorize as $q(\mathbf{x}) = \prod_k q_k(x_k)$. In other words, in problem (1) we approximate $p$ with a distribution that has independent variables. An approximation $q$ of this type is called mean field approximation (e.g., Jaakkola, 2000). Writing out in detail the right hand side of (1) one gets

$$F_{MF}(\{q_k\}) = -\int \log p(\mathbf{x}) \prod_k q_k(x_k) d\mathbf{x} + \sum_k \int q_k(x_k) \log q_k(x_k) dx_k.$$

Using notation $q_k(x_k) = N(x_k|m_i, \sigma_i^2)$, $\mathbf{m} = (m_1,\ldots,m_N)^T$ and $\boldsymbol{\sigma} = (\sigma_1,\ldots,\sigma_N)^T$, this simplifies to

$$F_{MF}(\mathbf{m}, \boldsymbol{\sigma}) = -\left\{\mathbf{h}^T\mathbf{m} - \frac{1}{2}\mathbf{m}^T\mathbf{J}\mathbf{m} - \frac{1}{2}\sum_k J_{kk}\sigma_k^2\right\} - \sum_k \log(\sigma_k) + C_{MF}, \quad (2)$$

where $C_{MF}$ is an irrelevant constant. Although $D\left[\prod_k q_k || p\right]$ might not be convex in $(q_1,\ldots,q_N)$, one can easily check that $F_{MF}$ is convex in its variables $\mathbf{m}$ and $\boldsymbol{\sigma}$ and its minimum is obtained for $\mathbf{m} = \mathbf{J}^{-1}\mathbf{h}$ and $\boldsymbol{\sigma} = 1/\sqrt{\operatorname{diag}(\mathbf{J})}$. Since $[\mathbf{J}^{-1}]_{kk} = 1/(J_{kk} - \mathbf{J}_{k,\backslash k}^T [\mathbf{J}_{\backslash k,\backslash k}]^{-1} \mathbf{J}_{\backslash k,k})$, one can easily see that the mean field approximation underestimates variances. Note that the mean field approximation computes a solution in which the means are exact, but the variances are computed as if there were no interactions between the variables, namely as if the matrix $\mathbf{J}$ were diagonal, thus giving poor estimates of the variances.

In order to improve the estimates for variances, one has to choose approximating distributions $q$ that are able to capture dependencies between the variables in $p$. It can be verified that any distribution in which the dependencies form a tree graph can be written in the form

$$p(\mathbf{x}) = \prod_{n(i,j)} \frac{p(x_i, x_j)}{p(x_i)p(x_j)} \prod_k p(x_k),$$

where $i$ and $j$ run through all the connections or edges $n(i,j)$ of the tree and $k$ runs through $\{1, \ldots, N\}$. Although in most cases the undirected graph generated by the connections in $\mathbf{J}$ is not a tree, based on the "tree intuition" one can construct $q$ from one and two variable marginals as

$$q(\mathbf{x}) \propto \prod_{n(i,j)} \frac{q_{ij}(x_i, x_j)}{q_i(x_i)q_j(x_j)} \prod_k q_k(x_k) \qquad (3)$$

and constrain the functions $q_{ij}$ and $q_k$ to be marginally consistent and normalize to 1, that is $\int q_{ij}(x_i, x_j) dx_j = q_i(x_i)$ for any $i$ and $j$ and $\int q_k(x_k) dx_k = 1$ for any $k$. An approximation of the form (3) together with the constraints on $q_{ij}$–s and $q_k$–s is called a Bethe approximation. Denoting the family of such functions by $\mathcal{F}_B$, by choosing $q_{ij}(x_i, x_j) = q_i(x_i)q_j(x_j)$ one can easily check that $\mathcal{F}_{MF} \subset \mathcal{F}_B$, thus $\mathcal{F}_B$ is non-empty. Assuming that the approximate marginals are correct and $q$ normalizes to 1 and then substituting (3) into (1), we get an approximation of the Kullback–Leibler divergence in (1) called the Bethe free energy. Since the interactions between the variables in $p$ are pairwise, we can factorize $p$ as $p(\mathbf{x}) \propto \prod_{n(i,j)} \Psi_{i,j}(x_i, x_j)$, and express the Bethe free energy as

$$F_B(\{q_{ij}, q_k\}) = -\sum_{n(i,j)} \int q_{ij}(\mathbf{x}_{i,j}) \log \Psi_{ij}(\mathbf{x}_{i,j}) d\mathbf{x}_{i,j}$$
$$+ \sum_{n(i,j)} \int q_{ij}(\mathbf{x}_{i,j}) \log \left[ \frac{q_{ij}(\mathbf{x}_{i,j})}{q_i(x_i)q_j(x_j)} \right] d\mathbf{x}_{i,j}$$
$$+ \sum_k \int q_k(x_k) \log q_k(x_k) dx_k. \qquad (4)$$

---

[1] The matrix $\mathbf{A}$ is diagonally dominant if $|A_{ii}| > \sum_{j \neq i} |A_{ij}|$ for all $i$.

[2] Following Malioutov et al. (2006), we call a Gaussian distribution pairwise normalizable if it can be factorized into a product of normalizable "pair" factors, that is $p(x_1,\ldots,x_n) = \prod_{ij} \Psi_{ij}(x_i, x_j)$ such that all $\Psi_{ij}$-s are normalizable.

Yedidia, Freeman, and Weiss (2000) showed that the fixed point iteration for finding the constrained minima of the function in (4), boils down to the message passing algorithm of Pearl (1988). The algorithm is derived from the Karush–Kuhn–Tucker conditions of the constrained minimization. As it was mentioned in the introduction, in case of Gaussian models this algorithm does not always converge, and the reason for this appears to be that the approximate marginals may get indefinite or negative definite covariance matrices. Welling and Teh (2001) pointed out that this can be due to the unboundedness of the Bethe free energy.

Since $F_{MF}$ is convex and bounded and the Bethe free energy could be unbounded, it seems plausible to analyze the fractional Bethe free energy

$$F_{\boldsymbol{\alpha}}(\{q_{ij}, q_k\}) = -\sum_{n(i,j)} \int q_{ij}(\mathbf{x}_{i,j}) \log \Psi_{ij}(\mathbf{x}_{i,j}) d\mathbf{x}_{i,j}$$
$$+ \sum_{n(i,j)} \frac{1}{\alpha_{ij}} \int q_{ij}(\mathbf{x}_{i,j}) \log \left[ \frac{q_{ij}(\mathbf{x}_{i,j})}{q_i(x_i)q_j(x_j)} \right] d\mathbf{x}_{i,j}$$
$$+ \sum_k \int q_k(x_k) \log q_k(x_k) dx_k \qquad (5)$$

introduced by Wiegerinck and Heskes (2003). Here, $\boldsymbol{\alpha}$ denotes the set of variables $\{\alpha_{ij}\}$. They showed that the fractional Bethe free energy "interpolates" between the mean field and the Bethe approximation. That is for $\alpha_{ij} = 1$ we get the Bethe free energy, while in the case when all $\alpha_{ij}$–s tend to 0 the mutual information between variables $x_i$ and $x_j$ is highly penalized, therefore, (5) enforces solutions close to the mean field solution. They also showed that the fractional message passing algorithm derived from (5) can be interpreted as Pearl's message passing algorithm with the difference that instead of computing local marginals—like in Pearl's algorithm—one computes local $\alpha_{ij}$–marginals[3]. The local $\alpha_{ij}$–marginals correspond to "true" local marginals when $\alpha_{ij} = 1$ and to local mean field approximations when $\alpha_{ij} = 0$.

Power expectation propagation by Minka (2004) is an approximate inference method that uses local approximations with $\alpha$–divergences. It turns out that in case of Gaussian models power expectation propagation—with a fully factorized approximating distribution—boils down to the same message passing algorithm as the one derived from (5) and the appropriate constraints.

Starting from the idea of creating a convex upper bound of the Bethe free energy when $p$ and $q$ are exponential distributions, Wainwright et al. (2003) derived a form of (5) where the $\alpha_{ij}$–s are chosen such that it is convex in $(\{q_{ij}\}, \{q_k\})$. Thus, they derived a form of the fractional Bethe free energy that has a unique global minimum.

## 3 Main results

In this section we analyze the parametric form of (5). Setting all $\alpha_{ij}$ values equal, we show that the fractional Gaussian Bethe free energy is a non-increasing function of $\alpha$. By letting $\alpha$ tend to infinity, we obtain a lower bound for the free energies. It turns out that the condition for the lower bound to be bounded from below is the same as the pairwise normalizability condition.

---

[3] Here, we define the $\alpha$–marginals of a distribution $p$ as $\mathrm{argmin}_{\{q_k\}} D_\alpha \left[ p \parallel \prod_k q_k \right]$, where $D_\alpha$ is the $\alpha$–divergence.

Conforming to Malioutov et al. (2006) and without loss of generality, we work with the "normalized" information matrix, that is we use $\mathbf{J} = \mathbf{I} + \mathbf{R}$ where $\text{diag}(\mathbf{R}) = \mathbf{0}$. We define $|\mathbf{R}|$ as the matrix formed by the absolute values of $\mathbf{R}$'s elements. Following Welling and Teh (2001), we use $q_{ij}(\mathbf{x}_{i,j}) = N(\mathbf{x}_{i,j}|\mathbf{m}_{ij}, \mathbf{\Sigma}_{ij})$ and $q_k(x_k) = N(x_k|m_k, \sigma_k^2)$, where $\mathbf{m} = (m_1, \ldots, m_N)^T$, $\mathbf{m}_{i,j} = (m_i, m_j)^T$ and $\mathbf{\Sigma}_{ij} = [\sigma_i^2, \sigma_{ij}; \sigma_{ij}, \sigma_j^2]$, and thus we embed the marginalization and normalization constraints into the parameterization. The matrix formed by diagonal elements $\sigma_k^2$ and off-diagonal elements $\sigma_{ij}$ is denoted by $\mathbf{\Sigma}$ and the vector of standard deviations by $\boldsymbol{\sigma} = (\sigma_1, \ldots, \sigma_N)^T$. Substituting $q_{ij}$ and $q_k$ into (5) one gets

$$F_{\boldsymbol{\alpha}}(\mathbf{m}, \mathbf{\Sigma}) = -\left\{\mathbf{h}^T\mathbf{m} - \frac{1}{2}\mathbf{m}^T\mathbf{J}\mathbf{m} - \frac{1}{2}\text{Tr}(\mathbf{J}^T\mathbf{\Sigma})\right\}$$
$$-\frac{1}{2}\sum_{n(i,j)}\frac{1}{\alpha_{ij}}\log\left(1 - \frac{\sigma_{ij}^2}{\sigma_i^2\sigma_j^2}\right)$$
$$-\sum_k \log(\sigma_k) + C_{\boldsymbol{\alpha}} \qquad (6)$$

where $C_{\boldsymbol{\alpha}}$ is an irrelevant constant. Note that the variables $\mathbf{m}$ and $\mathbf{\Sigma}$ are independent, hence the minimizations of $F_{\boldsymbol{\alpha}}(\mathbf{m}, \mathbf{\Sigma})$ with regard to $\mathbf{m}$ and $\mathbf{\Sigma}$ can be carried out independently.

**Property 1.** $F_{\boldsymbol{\alpha}}(\mathbf{m}, \mathbf{\Sigma})$ is convex and bounded in $(\mathbf{m}, \{\sigma_{ij}\}_{i\neq j})$ and at any stationary point we have

$$\mathbf{m}^* = \mathbf{J}^{-1}\mathbf{h} \qquad (7)$$
$$\sigma_{ij}^* = -\text{sign}(R_{ij})\frac{\left(1 + (2\alpha_{ij}R_{ij})^2\sigma_i^2\sigma_j^2\right)^{1/2} - 1}{2\alpha_{ij}|R_{ij}|}.$$

*Proof:* By definition $\mathbf{J}$ is positive definite, therefore, the quadratic term in $\mathbf{m}$ is convex and bounded. The variables $\mathbf{m}$ and $\mathbf{\Sigma}$ are independent and the minimum with regard to $\mathbf{m}$ is achieved at $\mathbf{m}^* = \mathbf{J}^{-1}\mathbf{h}$.

One can check that the second order derivative of $F_{\boldsymbol{\alpha}}(\mathbf{m}, \mathbf{\Sigma})$ with regard to $\sigma_{ij}$ is non-negative and the first order derivative has only one solution when $-\sigma_i\sigma_j \leq \sigma_{ij} \leq \sigma_i\sigma_j$ (Welling and Teh, 2001). Since the variables $\sigma_{ij}$ are independent, one can conclude that $F_{\boldsymbol{\alpha}}(\mathbf{m}, \mathbf{\Sigma})$ is convex in $\{\sigma_{ij}\}$. From the independence of $\mathbf{m}$ and $\mathbf{\Sigma}$, it follows that $F_{\boldsymbol{\alpha}}$ is convex in $(\mathbf{m}, \{\sigma_{ij}\})$. ∎

Since $\mathbf{\Sigma}_{ij}$ is constrained to be a covariance matrix, we have $\sigma_{ij}^2 \leq \sigma_i^2\sigma_j^2$, thus the first logarithmic term in (6) is negative. As a consequence, by setting all $\alpha_{ij}$-s equal, we get,

$$F_{\alpha_1}(\mathbf{m}, \mathbf{\Sigma}) \geq F_{\alpha_2}(\mathbf{m}, \mathbf{\Sigma}) \quad \text{for any } \alpha_1 \leq \alpha_2.$$

This leads to the following property.

**Property 2.** With $\alpha_{ij} = \alpha$, $F_\alpha$ is a non-increasing function of $\alpha$.

Using Property 1 and substituting $\sigma_{ij}^*$ into $F_{\boldsymbol{\alpha}}$ we define the constrained function

$$F_{\boldsymbol{\alpha}}^c(\mathbf{m}, \boldsymbol{\sigma}) = \frac{1}{2}\mathbf{m}\mathbf{J}^{-1}\mathbf{m} - \mathbf{h}^T\mathbf{m} + \frac{1}{2}\sum_k \sigma_k^2$$
$$-\frac{1}{2}\sum_{n(i,j)}\frac{1}{\alpha_{ij}}\left(\left(1 + (2\alpha_{ij}R_{ij})^2\sigma_i^2\sigma_j^2\right)^{1/2} - 1\right)$$
$$-\frac{1}{2}\sum_{n(i,j)}\frac{1}{\alpha_{ij}}\log\left(2\frac{\left(1 + (2\alpha_{ij}R_{ij})^2\sigma_i^2\sigma_j^2\right)^{1/2} - 1}{(2\alpha_{ij}R_{ij})^2\sigma_i^2\sigma_j^2}\right)$$
$$-\sum_k \log(\sigma_k) + C_{\boldsymbol{\alpha}}^c \qquad (8)$$

where $C_{\boldsymbol{\alpha}}^c$ is an irrelevant constant. From Property 2, it follows that when choosing $\alpha_{ij} = \alpha$, the function in (8) is a non-increasing function of $\alpha$. It then makes sense to take $\alpha \to \infty$ and verify whether we can get a lower bound for (8).

**Lemma** For any $\boldsymbol{\sigma} > 0$, $0 \leq \alpha_1 \leq 1$ and $\alpha_2 \geq 1$ the following inequalities hold.

$$F_{MF}(\mathbf{m}, \boldsymbol{\sigma}) \geq F_{\alpha_1}^c(\mathbf{m}, \boldsymbol{\sigma}) \geq F_B(\mathbf{m}, \{\sigma_{ij}^*\}, \boldsymbol{\sigma})$$
$$F_B(\mathbf{m}, \{\sigma_{ij}^*\}, \boldsymbol{\sigma}) \geq F_{\alpha_2}^c(\mathbf{m}, \boldsymbol{\sigma}) \ldots$$
$$\ldots \geq F_{MF}(\mathbf{m}, \boldsymbol{\sigma}) - \frac{1}{2}\boldsymbol{\sigma}^T|\mathbf{R}|\boldsymbol{\sigma}$$

Moreover, they are tight, that is

$$\lim_{\alpha \to 0} F_\alpha(\mathbf{m}, \{\sigma_{ij}^*(\alpha)\}, \boldsymbol{\sigma}) = F_{MF}(\mathbf{m}, \boldsymbol{\sigma})$$

and

$$\lim_{\alpha \to \infty} F_\alpha(\mathbf{m}, \{\sigma_{ij}^*(\alpha)\}, \boldsymbol{\sigma}) = F_{MF}(\mathbf{m}, \boldsymbol{\sigma}) - \frac{1}{2}\boldsymbol{\sigma}^T|\mathbf{R}|\boldsymbol{\sigma}.$$

*Proof:* Since the Bethe free energy is the specific case of the fractional Bethe free energy for $\alpha = 1$, the inequalities on $F_B(\mathbf{m}, \{\sigma_{ij}^*(\alpha)\}, \boldsymbol{\sigma})$ follow from Property 2. Now, we show that the upper and lower bounds are tight. The function $(1 + x^2)^{1/2} - 1$ behaves as $\frac{1}{2}x^2$ in the neighborhood of 0, therefore,

$$\lim_{\alpha \to 0} \sigma_{ij}^*(\alpha) = 0$$

and

$$\lim_{\alpha \to 0} \frac{\log\left(1 - \frac{\sigma_{ij}^{*\,2}(\alpha)}{\sigma_i^2\sigma_j^2}\right)}{\alpha} = -\frac{1}{\sigma_i^2\sigma_j^2}\lim_{\alpha \to 0}\frac{\sigma_{ij}^{*\,2}(\alpha)}{\alpha} = 0,$$

showing that $F_{MF}(\mathbf{m}, \boldsymbol{\sigma})$ is a tight upper bound. As $\alpha$ tends to infinity, we have

$$\lim_{\alpha \to \infty} \frac{\left(1 + (2\alpha R_{ij})^2 \sigma_i^2 \sigma_j^2\right)^{1/2} - 1}{2\alpha} = |R_{ij}|\sigma_i \sigma_j$$

and

$$\lim_{\alpha \to \infty} \frac{1}{\alpha} \log \left( \frac{\left(1 + (2\alpha R_{ij})^2 \sigma_i^2 \sigma_j^2\right)^{1/2} - 1}{(2\alpha R_{ij})^2 \sigma_i^2 \sigma_j^2} \right) = 0,$$

yielding

$$\lim_{\alpha \to \infty} F_\alpha(\mathbf{m}, \{\sigma_{ij}^*(\alpha)\}, \boldsymbol{\sigma}) = F_{MF}(\mathbf{m}, \boldsymbol{\sigma}) - \frac{1}{2}\boldsymbol{\sigma}^T |\mathbf{R}| \boldsymbol{\sigma}.$$

∎

Let $\lambda_{max}(|\mathbf{R}|)$ be the largest eigenvalue of $|\mathbf{R}|$. Analyzing the boundedness of the lower bound, we arrive at the following theorem.

**Theorem** *For the fractional Bethe free energy in (6) corresponding to a connected Gaussian network, the following statements hold*

(1) *if $\lambda_{max}(|\mathbf{R}|) < 1$, then $F_{\boldsymbol{\alpha}}$ is bounded from below for all $\boldsymbol{\alpha} > \mathbf{0}$,*

(2) *if $\lambda_{max}(|\mathbf{R}|) > 1$, then $F_{\boldsymbol{\alpha}}$ is unbounded from below for all $\boldsymbol{\alpha} > \mathbf{0}$,*

(3) *if $\lambda_{max}(|\mathbf{R}|) = 1$ then $F_{\boldsymbol{\alpha}}$ is bounded from below if and only if $\frac{1}{2} \sum_i \sum_{n(i,j)} \frac{1}{\alpha_{ij}} \geq N$.*

*Proof:* Since in $F_{\boldsymbol{\alpha}}$ there is no interaction between the parameters $\mathbf{m}$ and $\boldsymbol{\Sigma}$ and the term depending on $\mathbf{m}$ is bounded from below due to the positive definiteness of $\mathbf{J}$, we can simply neglect this term when analyzing the boundedness of $F_{\boldsymbol{\alpha}}$. Let us write out in detail the lower bound of the fractional Bethe free energies in the form

$$F_{MF}(\mathbf{m}, \boldsymbol{\sigma}) - \frac{1}{2}\boldsymbol{\sigma}^T |\mathbf{R}| \boldsymbol{\sigma} = \frac{1}{2}\mathbf{m}^T \mathbf{J}^{-1}\mathbf{m} - \mathbf{h}^T \mathbf{m}$$
$$+ \frac{1}{2}\boldsymbol{\sigma}^T (\mathbf{I} - |\mathbf{R}|) \boldsymbol{\sigma} - \sum_k \log(\sigma_k) + C, \quad (9)$$

*Statement (1):* The condition $\lambda_{max}(|\mathbf{R}|) < 1$ implies that $\mathbf{I} - |\mathbf{R}|$ is positive definite. Now, $\log(x) \leq x - 1$, thus $\frac{1}{2}\boldsymbol{\sigma}^T(\mathbf{I} - |\mathbf{R}|)\boldsymbol{\sigma} - \mathbf{1}^T \log(\boldsymbol{\sigma}) \geq \frac{1}{2}\boldsymbol{\sigma}^T(\mathbf{I} - |\mathbf{R}|)\boldsymbol{\sigma} - \mathbf{1}^T\boldsymbol{\sigma} + N$. The latter is bounded from below and so it follows that (9) is bounded from below as well. According to the Lemma, boundedness of (9) implies that all fractional Bethe free energies are bounded from below.

*Statement (2):* Since we assumed that the Gaussian network is connected and undirected, it follows that $|\mathbf{R}|$ is irreducible (e.g., Horn and Johnson, 2005). According to the Frobenius-Perron theory of non-negative matrices (e.g., Horn and Johnson, 2005), the non-negative and irreducible matrix $|\mathbf{R}|$ has a simple maximal eigenvalue $\lambda_{max}(|\mathbf{R}|)$ and all elements of the eigenvector $\mathbf{u}_{max}$ corresponding to it are positive. Let us take the fractional Bethe free energy and analyze its behavior when $\boldsymbol{\sigma} = t\mathbf{u}_{max}$ and $t \to \infty$. Since for large values of $t$ we have $\left(1 + (2\alpha_{ij}R_{ij})^2(u_{max}^i u_{max}^j)^2 t^4\right)^{1/2} \simeq 2\alpha_{ij}|R_{ij}|u_{max}^i u_{max}^j t^2$, the sum of the second and third term in (8) boils down to $(1 - \lambda_{max}(|\mathbf{R}|))t^2$ and this term dominates over the logarithmic ones as $t \to \infty$. As a result, the limit is independent of the choice of $\alpha_{ij}$ and it tends to $-\infty$ whenever $\lambda_{max}(|\mathbf{R}|) > 1$.

*Statement (3):* If $\lambda_{max}(|\mathbf{R}|) = 1$, then the only direction in which the quadratic term will not dominate is $\boldsymbol{\sigma} = t\mathbf{u}_{max}$. Therefore, we have to analyze the behavior of the logarithmic terms in (8) when $t \to \infty$. For large $t$-s these terms behave as $\left(\frac{1}{2}\sum_i \sum_{n(i,j)} \frac{1}{\alpha_{ij}} - N\right) \log(t)$. For this reason, the boundedness of $F_{\boldsymbol{\alpha}}^c$—and thus of $F_{\boldsymbol{\alpha}}$—depends on the condition in statement (3). ∎

It was shown by Malioutov et al. (2006) that the condition $\lambda_{max}(|\mathbf{R}|) < 1$ is an equivalent condition of pairwise normalizability. Therefore, pairwise normalizability is not only a sufficient condition for the message passing algorithm to converge, but it is also a necessary condition for the fractional Gaussian Bethe free energies to be bounded.

**Example** In the case of models with K–regular adjacency matrix (non-zero entries of $\mathbf{R}$) and equal interaction weights $R_{ij} = r$, the maximal eigenvalue of $|\mathbf{R}|$ is $\lambda_{max}(|\mathbf{R}|) = Kr$ and the eigenvector corresponding to this eigenvalue is $\mathbf{1}$. (We define $\mathbf{1}$ as the vector that has all its elements equal to 1.) Verifying the stationary point conditions, it turns out that for some choice of $r$ and $\alpha$ there exists a local minimum which is symmetrical, that is it lies in the direction $\mathbf{1}$. One can show that when the model is not pairwise normalizable ($Kr > 1$), the critical $r$ below which the fractional Bethe free energy possesses this local minimum is $r_c(K, \alpha) = 1/2\sqrt{\alpha(K - \alpha)}$ and for any valid $r$ the critical $\alpha$ below which the fractional Bethe free energies possesses this local minimum is $\alpha_c(K, r) = \frac{1}{2}K\left(1 - \sqrt{1 - 1/(Kr)^2}\right)$. These results are illustrated in Figure 1. (Note that for 2–regular graphs, all valid models are pairwise normalizable and possess a unique global minimum.) ∎

For K–regular graphs convexity of the fractional Bethe free energy in terms of $\{q_{ij}, q_k\}$ requires $\alpha \geq K$, a much stronger condition than $\alpha \geq \alpha_c(K, r)$. Thus, if we choose $\alpha$ sufficiently large such that the Bethe free energy is guaranteed to have a unique global minimum, this minimum is unbounded.

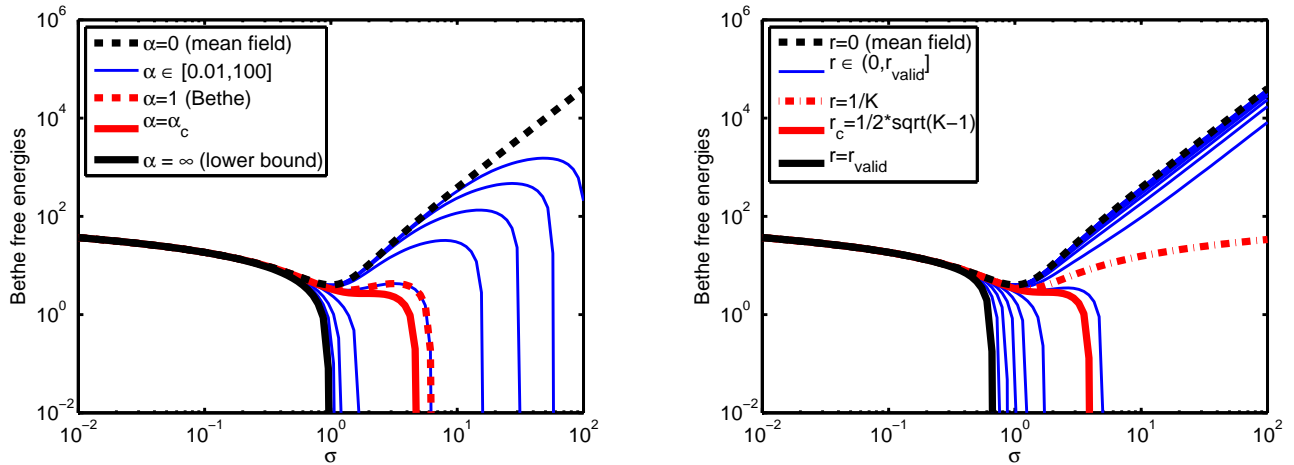

Figure 1: Visualizing critical parameters for a symmetric K-regular Gaussian model with $R_{ij} = r$. Plots in the left panel correspond to the constrained fractional Bethe free energies $F_\alpha^c$ in the direction $\mathbf{1}$ for an 8 node 4–regular Gaussian model with r=0.27 ($Kr > 1$) and varying $\alpha$. Plots in the right panel correspond to the constrained Bethe free energies $F_1^c$ in the direction $\mathbf{1}$ for an 8 node 4–regular Gaussian model with varying $r$. Here, $r_{valid}$ is the supremum of $r$–s for which the model is valid—$\mathbf{J}$ is positive definite.

## 4 Experiments

We implemented both direct minimization and fractional message passing and analyzed their behavior for different values of $\lambda_{max}(|\mathbf{R}|)$. For reasons of simplicity we set all $\alpha_{ij}$ equal. The results are summarized in Figure 2. Note that there is a good correspondence between the behavior of the fractional Bethe free energies in the direction of the eigenvalue corresponding to $\lambda_{max}(|\mathbf{R}|)$ and the convergence of the Newton method. The Newton method was started from varying initial points. We experienced that when $\lambda_{max}(|\mathbf{R}|) > 1$ and setting the initial value to $t\mathbf{u}_{max}$, the algorithm did not converge for high values of $t$. This can be explained by the plots in Figure 2: for high values of $t$ the initial point is not in the convergence region of the local minimum. For the fractional message passing algorithm we used two types of initialization: (1) when $\lambda_{max}(|\mathbf{R}|) < 1$ we set $\Psi_{ij}$ such that they are all normalizable and set initial messages to 1, (2) when $\lambda_{max}(|\mathbf{R}|) \geq 1$, we used a symmetric partitioning of $p$ into $\Psi_{ij}$-s —symmetric partitioning of the diagonal elements of $\mathbf{J}$—and set messages to identical values such that in the first step of the algorithm all approximate marginals become normalizable.

We experienced a behavior similar to that described by Welling and Teh (2001) for standard message passing, namely fractional message passing and direct minimization either both converge or both fail to converge. Our experiments in combination with the Theorem show that when $\lambda_{max}(|\mathbf{R}|) > 1$ standard message passing at best converges to a local minimum of the Bethe free energy. If standard message passing fails to converge, one can decrease $\alpha$ and search for a stationary point—preferably a local minimum—of the corresponding fractional free energy.

It can be seen from the results in the right panels of Figure 1 and 2, that when the model is no longer pairwise normalizable, the local minimum and not the unbounded global minimum is the natural continuation of the (bounded) global minimum for pairwise normalizable models. This explains why the quality of the approximation at the local minimum for models that are not pairwise normalizable is still comparable to that at the global minimum for models that are pairwise normalizable. Note that when the model is pairwise normalizable both the upper and lower bounds are convex in $t$. This may be obscured by the use of logarithmic axes.

## 5 Conclusions and future research

As we have seen, $F_{MF}$ and $F_{MF} - \frac{1}{2}\boldsymbol{\sigma}^T|\mathbf{R}|\boldsymbol{\sigma}$ provide tight upper and lower bounds for the Gaussian fractional Bethe free energies. It turns out that pairwise normalizability (see Malioutov et al., 2006) is not only a sufficient condition for the message passing algorithm to converge, but it is also a necessary condition for the

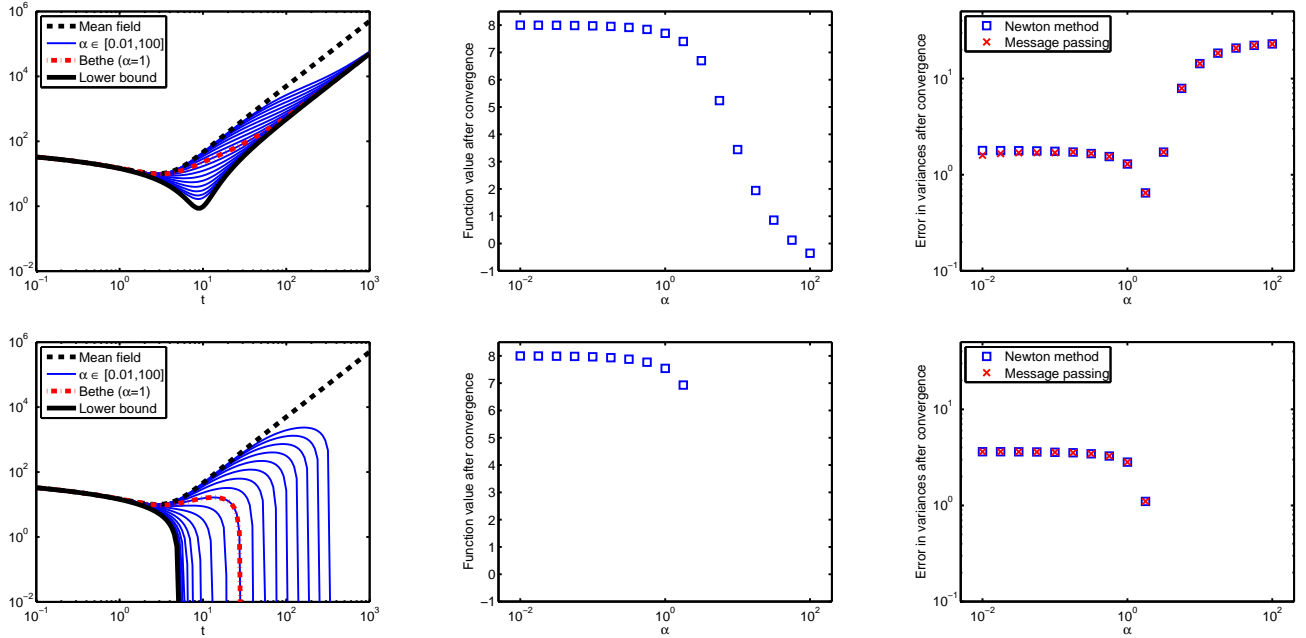

Figure 2: Left panels show the constrained fractional Bethe free energies of an 8 node Gaussian network in the direction of the eigenvector corresponding to $\lambda_{max}(|\mathbf{R}|)$ for $\lambda_{max}(|\mathbf{R}|) = 0.9$ (top) and $\lambda_{max}(|\mathbf{R}|) = 1.1$ (bottom). The thick lines are the functions $F_{MF}$ (dashed), $F_B$ (dashed dotted) and the lower bound $F_{MF} - \frac{1}{2}\sigma^T|\mathbf{R}|\sigma$ (continuous). The thin lines are the constrained $\alpha$-fractional free energies $F_\alpha^c$ for $\alpha \in [10^{-2}, 10^2]$. Center panels show the final function values after the convergence of the Newton method. Right panels show the $||\cdot||_2$ error in approximation for the single node standard deviations $\boldsymbol{\sigma}$. Missing values indicate non-convergence.

Gaussian fractional Bethe free energies to be bounded from below.

If the model is pairwise normalizable, then the lower bound is bounded from below, therefore, both direct minimization and fractional message passing are converging for any choice of $\boldsymbol{\alpha} > \mathbf{0}$. Experiments show that regardless of the initialization, they converge to the same minimum. This suggests that in the pairwise normalizable case, fractional Bethe free energies possess a unique global minimum.

If the model is not pairwise normalizable, then none of the fractional Bethe free energies is bounded from below. However, experiments show that there is always a range of $\alpha$ values for which both direct minimization and fractional message passing converge. By decreasing $\alpha$ towards zero, one gets closer to the mean field energy and a local minimum to which the minimization algorithms can converge may appear.

As mentioned in Section 2, $\alpha_{ij}$-s correspond to using local $\alpha_{ij}$ divergences when applying power expectation propagation with a fully factorized approximating distribution. Several papers on continuous models (e.g., Minka, 2004; Qi et al., 2005) report that when expectation propagation does not converge, applying power expectation propagation with $\alpha < 1$ helps to achieve convergence. In the case of the problem addressed in this paper this behavior can be explained by the observation that small $\alpha$-s make local minima more likely to occur and thus prevents the covariance matrices from becoming indefinite or even non positive definite.

Wainwright et al. (2003) propose to convexify the Bethe free energy by choosing $\alpha_{ij}$-s sufficiently large such that the fractional Bethe free energy has a unique global minimum. This strategy appears to fail for Gaussian models. Convexification makes the possibly useful local minima dissapear, leaving just the unbounded global minimum. In the case of the more general hybrid models the use of the convexification is still unclear.

Note that the example in the previous section disproves the conjecture in Welling and Teh (2001): even when the Bethe free energy is not bounded from below, it can possess a local minimum to which the minimization algorithms converge.

Our future goals are to find sufficient conditions for the Bethe free energy to possess local minima and to derive algorithms that are guaranteed to find those.

**Acknowledgements**

We would like to thank Jason K. Johnson for suggesting us to try a case study of models with K–regular adjacency matrix and equal interaction weights, and sharing his ideas about the the walk–sum analysis of these models. The authors would like to acknowledge the support from the Vici grant 639.023.604 (second author) and Marie Curie EST program MEST-CT-2004-514510 (first author).